# Bistatic SAR ATR

A.K. Mishra and B. Mulgrew

**Abstract:** With the present revival of interest in bistatic radar systems, research in that area has gained momentum. Given some of the strategic advantages for a bistatic configuration, and technological advances in the past few years, large-scale implementation of the bistatic systems is a scope for the near future. If the bistatic systems are to replace the monostatic systems (at least partially), then all the existing usages of a monostatic system should be manageable in a bistatic system. A detailed investigation of the possibilities of an automatic target recognition (ATR) facility in a bistatic radar system is presented. Because of the lack of data, experiments were carried out on simulated data. Still, the results are positive and make a positive case for the introduction of the bistatic configuration. First, it was found that, contrary to the popular expectation that the bistatic ATR performance might be substantially worse than the monostatic ATR performance, the bistatic ATR performed fairly well (though not better than the monostatic ATR). Second, the ATR performance does not deteriorate substantially with increasing bistatic angle. Last, the polarimetric data from bistatic scattering were found to have distinct information, contrary to expert opinions. Along with these results, suggestions were also made about how to stabilise the bistatic-ATR performance with changing bistatic angle. Finally, a new fast and robust ATR algorithm (developed in the present work) has been presented.

## 1 Introduction

As per the standard definition [1], the original radar systems were bistatic, as they had physically separated transmitters and receivers. Soon, monostatic radars replaced them, because of the ease of design and compactness. Recently, there has been a revived interest in bistatic radars. This is because of a range of strategic advantages of a bistatic configuration as compared with its monostatic counterpart [2]. Second, a bistatic configuration is the first step towards a multistatic configuration, which may confer additional operational and strategic advantages like the netted radar system [3].

From the user point of view, replacing an existing monostatic system should be justified by an equally apt radar system, which can handle the existing usages and offer other added advantages. One such usage of primary importance in most of the existing airborne (monostatic) radar systems is the facility of automatic target recognition (ATR) (to various degrees of autonomy). In the recent years some studies have come out in the open literature, describing object recognition using passive radars [4, 5]. Passive radars are, strictly speaking, bistatic radars. However, there has been no work in the open literature reporting ATR algorithms for airborne bistatic synthetic aperture radar (SAR)-based ATR. In the present work, we have dealt with this problem of bistatic ATR using airborne SAR images. There are quite a few novelties involved in this work. First of all, a study of bistatic ATR in itself is of huge practical importance and is a novel work in itself.

Looking at the present level of bistatic radar development, there was a lack of database to carry out these ATR exercises. The second novelty of the current work involves the development of the setup for the generation of synthetic bistatic SAR-signature database.

As the data were generated synthetically, the data of all different combinations of polarisations were simulated. In looking for the information content in the bistatic multipolar data, it was found that the bistatic multipolar data contain information about the target, in contrast to what has been predicted by some researchers [6].

Finally, a new algorithm for ATR has been developed, based on a principal component analysis (PCA)-based nearest neighbour (NN) approach, which being simple and extremely fast, is expected to be of practical importance [7].

The paper is arranged as follows. Section 2 consists of a brief overview of a generic SAR ATR system. This is followed by a discussion of the challenges involved in developing bistatic SAR ATR, in Section 3. In Section 4, we discuss the algorithms used in the present work, and the parameters on which to compare the monostatic and the bistatic ATR results. The next section deals with the basis on which we looked for information in bistatic multipolar data. Finally, we present the results. This is followed by the section where discussions on the limitations of the present study are presented. The paper ends with a discussion of the major conclusions.

## 2 SAR ATR: an overview

There has been a great deal of research in the field of radar-based ATR systems. This is mainly because of the all-weather and covert abilities of a radar system, and the increasing image-resolution possible using SAR-imaging. The complexities in this field are numerous. The way a radar image is generated makes SAR image recognition problem different from the general optical image recognition problem. The process of SAR image formation of a target or a scene is a sensitive function of the range of the system and scene parameters. Hence, the ATR algorithms need to be extremely robust to any variation in the imaging system and scene parameters. To add to the complexity, the strategic nature of the usage demands a much lower acceptable error-in-classification level than that is acceptable in many other classification exercises. There is also the problem of getting decent training data, which can be both expensive and time consuming. There are a few reports handling end-to-end SAR ATR exercises in the open literature [8, 9], and many other diverse approaches towards handling exclusively the task of target recognition. Some of the reports with excellent ATR performance reported have used a range of classification algorithms, starting from simple template matching [9] and Gaussian-modelled Bayesian approach [10] to those involving more involved algorithms such as the support vector machine approaches [11].

## 3 Bistatic SAR ATR experiments: challenges and answers

One of the basic requirements of any target classification exercise is the database of target signatures, on which the algorithms are to be tested. For the stand-alone problem of monostatic ATR-algorithm development, there is a widely accepted archive of SAR images of a wide variety of targets [12]. Unlike the monostatic counterpart, in the bistatic case there are no datasets in the public domain, which could be used in validation and analysis of any classification algorithm. This was a major challenge for the present work. Even though there are some field-generated bistatic SAR images, they were not exhaustive enough to be used in an ATR exercise. As the next best alternative, an electro magnetic modelling tool [13] was used to model targets and to generate a database of bistatic SAR image clips (A target clip is the SAR image of the target, with the image of the target at the centre of the image. In an automatic target recognition exercise, after the detection stage, a particular part of the original SAR image of the scene is taken for further processing. This part of the SAR image, with the target at its centre, is termed as the target-clip. In the present work, only the classification problem is handled. Hence, the inputs taken are the target image clips. Though no clipping operation is done, for convention, the word target clip is used through out.) [14].

In modelling the targets (military land vehicles), only the major (classifiable) features were modelled, ignoring the finer details. This approach served two main purposes: (i) it made the modelling job simpler; (ii) it also tallied with the requirement of any robust ATR-exercise, viz. the algorithm should not depend on the finer details of a target (the ones that are most likely to change a lot). On the basis of this principle of modelling major features, four generic targets were modelled for simulation: a main battle tank (MBT), an armoured personnel carrier (APC), a stinger missile launcher (STR) and a land missile launcher (MSL).

The collection of radar data was accomplished as follows. The model was kept static and the simulated transmitter was kept fixed at a point. The simulated receiver was rotated round the target at a fixed elevation, and at the same time not exceeding a certain particular maximum value of bistatic angle. Fig. 1 gives the geometry of data-collection simulation environment. Radar returns were collected from the scene with a target.

For a given position of transmitter and receiver, a range of frequencies were transmitted, the response to which forms the range resolution profile of the target at the given positions of the transmitter and the receiver. Collection of the range profiles for a range of positions of transmitter and receiver forms a sampled space from the $k$-space for the scene to be imaged. Patches from the $k$-space were taken to form the SAR image clips of the target [14–16]. Range profiles were collected for 24 different positions of the transmitter so as to gather the information of the targets from different aspect angles. The transmitter positions for which the data were collected are given below in Table 1. The number of transmitter positions was limited primarily by the simulation time. While forming the images, uniform patches were collected from the $k$-space data. Hence the resolutions of the images are variable depending on the bistatic angle of data collection. The best resolution attained in the database was for minimum bistatic angle, which was around 0.3 m in both dimensions. For this the bandwidth used was around 450 MHz.

In Fig. 2, the target models (not to scale) are illustrated and in Fig. 3, the corresponding SAR images are shown. The SAR images are one of the realisations, with minimum bistatic angle.

The second challenge for a project like this, where one needs a comparative study of ATR in the monostatic and the bistatic scenario, is the generation of a monostatic database as well. Because, a fair comparison is possible only when the databases are similar in the way they are generated. In the monostatic simulation, the simulated radar platform (with both transmitter and receiver) was moved round the target at a fixed elevation angle, collecting the radar returns from the target for a range of frequencies. This in turn formed the $k$-space data, patches from which were postprocessed to generate the SAR image clips.

The final database had 700 image clips for each target in the training dataset, and 700 image clips for each target in the test dataset.

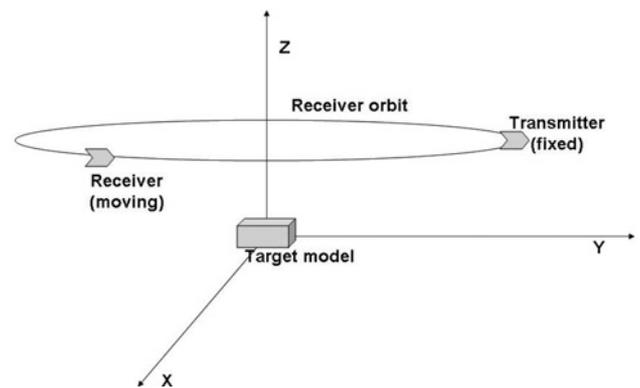

**Fig. 1** *Geometry of the simulated data-collection platforms*

**Table 1:** Elevation and azimuth angle combinations of the transmitter for which data were collected

| Elevation (in deg.) | Azimuth (in deg.) |
|---|---|
| 10 | 0, 60, 120, 180, 240, 300 |
| 15 | 0, 60, 120, 180, 240, 300 |

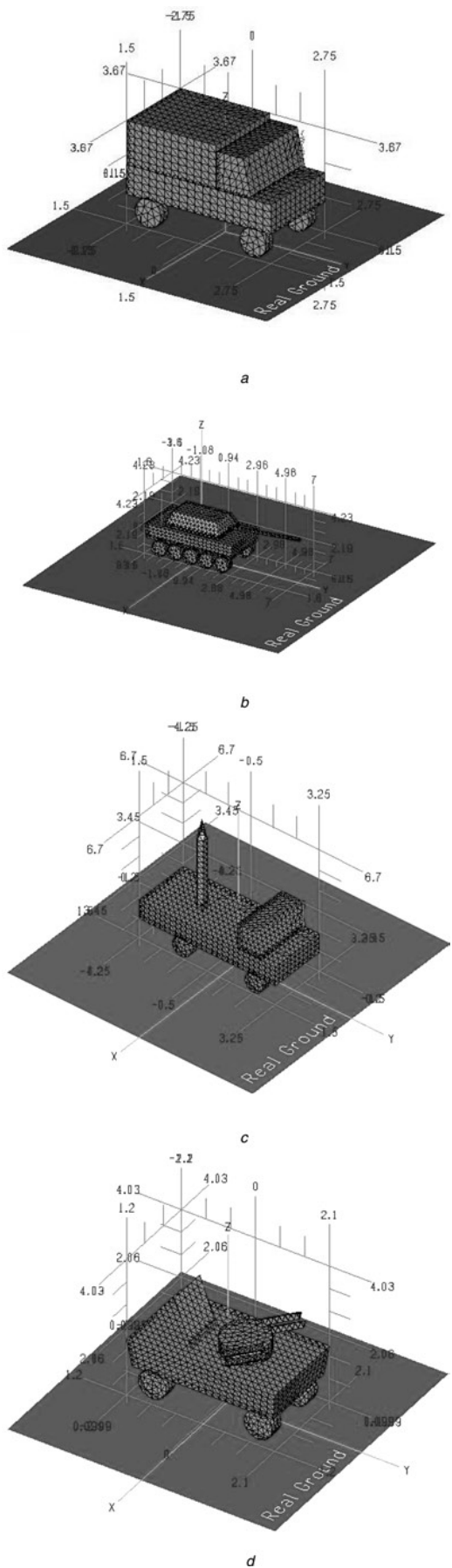

**Fig. 2** *CAD models of the targets simulated in the present work (dimensions in metres)*

*a* Armoured personnel carrier
*b* Main battle tank
*c* Missile launcher
*d* Stringer launcher

## 4 ATR algorithms used

In the present work, two algorithms were used for the ATR exercises. In both the algorithms, the database is arranged so that all image clips collected at 10° of receiver elevation are taken as the training dataset, and those collected at 15° of receiver elevation are taken as the test dataset.

### 4.1 Conditional Gaussian-model-based Bayesian classifier (CGBC)

The conditional Gaussian-model-based Bayesian classifier is an efficient and nearly optimum classifier because of its closeness to a Bayesian classifier. This has been confirmed by the excellent results reported using this method for monostatic ATR [10]. In this algorithm each pixel of the image clip is assumed to follow Gaussian distribution, depending on the target type and the receiver azimuth (For monostatic case, it is the target pose, but in bistatic case, receiver azimuth was found to be a more simple and convenient parameter).

$$r = s(\theta, a) + w \qquad (1)$$

Here, $r$ is the observed intensity of a pixel from the image, $w$ the Gaussian noise, and $s$ the complex-valued signal depending upon $\theta$ (the receiver azimuth angle) and $a$ (the target type). The log-likelihood of an observed $r$, given $\{\theta, a\}$ can be shown to be proportional to [10]

$$-\sum_{i=1}^{N}\left[\log(\sigma_i) + \left(\frac{r_i - \mu_i}{\sigma_i}\right)^2\right] \qquad (2)$$

Here $r_i$ is the $i$th pixel amplitude of the test-image-clip, $\sigma_i$, $\mu_i$ are the variance and the mean of the $i$th pixel, respectively (as estimated from the training data), and $N$ is the total number of pixels in the image-clip. In this method, the recognition is done as per the Bayesian rule of maximising the probability

$$P(a|r) = P(r|a)P(a) \qquad (3)$$

Here, $P(a)$ is the probability for the detection of each type of vehicle, and is taken to be equal in the current work.

### 4.2 Principal component analysis based nearest neighbour (PCANN) algorithm

The other ATR algorithm used in the present work has been developed in the present work. This algorithm is a feature-based classifier, where the features are the approximate principal components of the radar image data. PCA is performed on the SAR image data to extract the features, and the simple NN algorithm is used to handle the main classification task. It has been coined as the PCANN algorithm [7]. In this PCA-based [17, 18] algorithm, the image pixels are assumed to be the observed variables, depending upon the target type.

$$r = s(a) + w \qquad (4)$$

Here, $r$ is the observed intensity of a pixel from the image, $w$ Gaussian noise and $s$ the complex-valued signal depending upon $a$ (the target type). Pixels of an image clip are assumed to take different observational values with different image clips of the same class (which are collected with the same receiver elevation angle, but different receiver azimuth angles). PCA is applied to the dataset to reduce the number of observed variables. The complete ATR process can be summarised in the following steps:

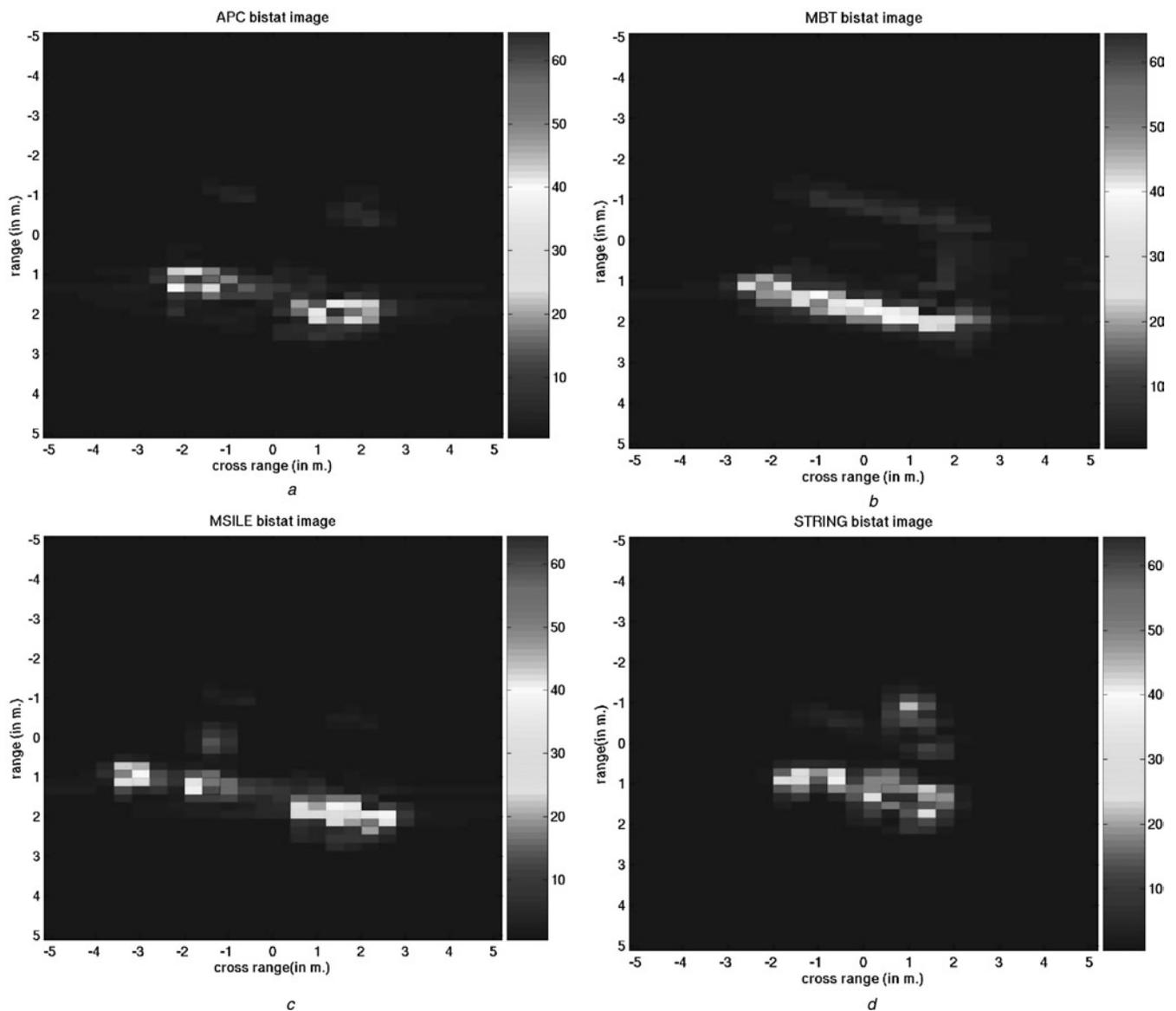

**Fig. 3** *Bistatic SAR images of the targets modelled (average bistatic angle of imaging is 15°)*
*a* Armoured personnel carrier
*b* Main battle tank
*c* Missile launcher
*d* Stringer launcher

- For each image clip, the pixels are stacked into a vector, and the consecutive image clips are taken as different observational vectors.
- All consecutive image-pixel vectors are arranged together to form the observation matrix. In the observation matrix, each column corresponds to an image clip and consecutive columns are formed out of consecutive image-clips from the training dataset.
- The observation matrix is normalised (to have unity variance), and all the observation vectors are made zero centred (by removing the mean). Let the resulting observation matrix be denoted by $X$.
- From this observation matrix, the covariance matrix is found for the observation vector.

$$Q = X^H X \quad (5)$$

- Then the eigenvalue operation is applied on $Q$ to get the eigen vectors.
- Eigenvectors corresponding to $n$ largest eigenvalues are stacked together to form matrix $V$. ($n$ is to be determined from the dataset, to represent maximum amount of variance present in the dataset. For all the present exercises, $n = 20$ was found to be an optimum choice.)
- Using this matrix $V$, the training dataset is reduced in dimension to $n$. The final output from the training phase is the database in reduced dimension and the converting matrix $V$.
- In the test phase, the test image clip is reduced in dimension using the converting matrix $V$.
- Next, the Euclidean distance is found from each point in the training database. The class of target, giving the least distance, is decided as the class of the test image clip.

There are two major advantages in the PCANN algorithm. First of all, as it handles the data in the PC-domain (which is of much reduced dimension than the original dataset), computationally the algorithm is extremely fast. Second, as shown in [19], the data extracted by the PCA analysis of SAR images closely correspond to the information obtained by scattering centre analysis. Scattering centre analysis has been proved to be a powerful method in SAR ATR [20]. Hence, the PCANN ATR algorithm can give most of the advantages of a scattering-centre-based ATR approach, at a much lower computational cost. An analysis showing the correspondence between PCA and the scattering centre analysis of radar data is given in the Appendix 1. Further it can be mentioned here that 20 principal components have been used in all the exercises reported in the current paper. This was because of two reasons. First, PCA of the SAR images generated in the current project showed that around 20 principal components will account for 99% of the data-variance. This is an accepted way of determining the number of PCs to be used for a particular task [17].

### 4.3 Parameters of comparison

The major thrust of the present work is the comparison of the relative performance of the bistatic ATR and the monostatic ATR. For detailed analysis of the PCANN algorithm, reader s are requested to refer to one of our earlier publications [21], which explains PCANN ATR algorithm as applied to monostatic SAR ATR. In another publication of ours [22], the PCANN ATR algorithm has been analysed for its advantages and use for bistatic SAR ATR. However, for the sake of continuity, the receiver output characteristic (ROC) curves for the two algorithms, CGBC and PCANN, will be presented.

In performing the comparison, the following questions will be addressed. These are some of the issues of immense practical importance, given the ever impending question of the choice between monostatic and bistatic systems.

- Can the performance of the bistatic ATR match the performance obtained from a monostatic system?
- What is the figure of deterioration in the ATR performance, for an increase in the bistatic angle of operation?
- If there is any loss of performance, can it be made up for?

## 5 Information from bistatic fully polarised data

In the present work, a synthetic database has been used. Hence, the fully polarimetric data were available (for the bistatic configuration). A limited study was undertaken to examine if there is any possible information from the fully polarimetric data in bistatic configuration and the use of the same for improving the bistatic SAR ATR.

The fully polarimetric radar systems have been the subject of intense study and research for the past few decades. It has been shown that the fully polarimetric data in monostatic configuration give a lot of information about the physical features of the target. Especially remarkable is the work by Huynen [6], which gave a one-to-one correspondence between the physical features of a target and the parameters derived from the fully polarimetric data collected using a monostatic radar system. However, because of the large number of combinations of transmitter and receiver, involved in bistatic SAR imaging, Huynen has expressed strict reservation against any information content in the bistatic fully polarimetric data [6].

In the rest of this section some of the basics of radar polarimetry will be described. In the right-hand Cartesian coordinate system ($x, y, z$), if the $z$-direction is taken as the direction of propagation for the electromagnetic wave, the field (For simplicity we deal only with the electrical wave in this work, though similar treatment could be done taking the magnetic wave instead.) can be represented as

$$\boldsymbol{E} = \hat{\boldsymbol{u}}_x E_x + \hat{\boldsymbol{u}}_y E_y \quad (6)$$

Here $\hat{\boldsymbol{u}}_x$ and $\hat{\boldsymbol{u}}_y$ are unit vectors (vectors of unit amplitude) in the $x$ and the $y$ directions. In the simplest horizontal and vertical (HV) basis, this expression becomes

$$\boldsymbol{E} = \hat{\boldsymbol{u}}_h E_h + \hat{\boldsymbol{u}}_v E_v \quad (7)$$

In matrix format this is represented by the Jones vector.

$$\boldsymbol{E} = \begin{bmatrix} E_h \\ E_v \end{bmatrix} \quad (8)$$

To make the data real valued and to attach some geometrical representation to this, the preferred way to represent a polarised electromagnetic wave is the Stokes vector, $\boldsymbol{g}(E)$.

$$\boldsymbol{g}(E) = \frac{1}{\sqrt{|E_h|^2 + |E_v|^2}} \begin{bmatrix} |E_h|^2 + |E_v|^2 \\ |E_h|^2 - |E_v|^2 \\ 2\,\text{Re}\,\{E_h^* E_v\} \\ 2\,\text{Im}\,\{E_h^* E_v\} \end{bmatrix} \quad (9)$$

The matrix relating the scattered and transmitted Jones vector is called the scattering or Sinclair matrix. Similarly the matrix relating the Stokes vectors for the transmitted, and the scattered wave is the Kennaugh matrix or the $\boldsymbol{K}$ matrix (which is called the Muller's matrix $\boldsymbol{M}$, for the forward scattering case).

$$\boldsymbol{g}(E^s) = [\boldsymbol{K}]\boldsymbol{g}(E^t) \quad (10)$$

Here, $\boldsymbol{E}^s$ and $\boldsymbol{E}^t$ represent the scattered and the transmitted electric fields, respectively. Elements of the $\boldsymbol{K}$-matrix have been linked to the Huynen's parameters. These parameters are represented by $A_0, B_0, B, C, D, E, F, G$ and $H$. For the monostatic case, these parameters have been shown to have direct relation to the physical features of the target [6]. For the monostatic case, the $\boldsymbol{K}$ matrix is symmetric, whereas in bistatic case it is not. Hence, the number of independent parameters extracted from the $\boldsymbol{K}$ matrix is more in a bistatic case. In one of their works, Germond et al. [23] have shown that there are seven more independent parameters which can be derived from the $\boldsymbol{K}$ matrix for a bistatic case.

$$[\boldsymbol{K}_{\text{bi}}] = \begin{bmatrix} A_0 + B_0 + A & C + I \\ C - I & A_0 + B - A \\ H - N & E - K \\ F - L & G - M \\ H + N & F + L \\ E + K & G + M \\ A_0 - B - A & D + J \\ D - J & B_0 - A_0 - A \end{bmatrix} \quad (11)$$

Here $[\boldsymbol{K}_{\text{bi}}]$ represents the $\boldsymbol{K}$ matrix for the bistatic case, and $A, I, J, K, L, M$ and $N$ are the extra seven parameters for the bistatic case. Germond et al. [23] have given closed-form expressions to find them using the elements of the scattering matrix. However, Huynen has strongly stated that any such bistatic parameters would contain no information about the target [6].

In the current work, to look for the information content in the parameters derived from the $\boldsymbol{K}$-matrix, those parameters were extracted for each pixel of the SAR image, using the corresponding SAR images from all the four combinations of polarisations (HH, HV, VH and VV) [24]. These derived parameters were in turn used as features of the target, for ATR. Two basic assumptions made in this exercise were:

1. The complex amplitudes of a particular pixel from the HH, HV, VH and VV polarised images give the approximate elements for the Sinclair or the scattering matrix corresponding to the scattering element represented by that pixel. This assumption is a common one used by most of the researchers trying to apply polarimetry to the field of ATR [24].
2. Second, it was assumed that if a certain derived parameter derived from the $K$-matrix has any physical significance, then adding that parameter as a feature should increase the ATR performance.

The algorithm used for ATR is that of multi dimensional PCANN [21] neighbour classifier. The initial SAR image can be modelled as a two-dimensional matrix. After generating the parameters from the $K$-matrix for each pixel (from the four polarised images), it results in a set of two-dimensional matrices. For example, if $k$ parameters are to be considered, then for each pixel there are extracted $k$ parameters. This results in $k$ two-dimensional matrices. The complete data can now be treated as a three-dimensional matrix. Then PCA is applied to reduce the dimensions and the correlation in the data. Because this needs applying PCA to each dimension of the data, this is called multidimensional PCA. The resulting data are much reduced in size (For example in the current exercises, the image clips were of size $50 \times 50$. Each such frame is reduced to 20 PCs. Hence, if 8 parameters are extracted for each pixel, the original data would be $50 \times 50 \times 8 = 20\,000$ data points. By applying PCA, the 8 parameters are reduced to 2, and hence the final data size after a 3D PCA is just $20 \times 2 = 40$ data points.) and is used in turn in ATR, using the NN recognition algorithm.

## 6 Results and observations

### 6.1 Comparing the ATR algorithms

The ROCs for bistatic ATR for the two algorithms are given in Fig. 4. For the ease of comparison, separate sets of ROC curves have been drawn for the four types of targets modelled. It can be observed that the ROC curves are not much worse than the ROC curves found for the case of monostatic ATR, in the open literature [10]. Hence, bistatic ATR can give performance almost as good as a monostatic ATR system. As far as comparison of the PCANN ATR algorithm with the standard CGBC algorithm is concerned, it can be observed that except for the target APC, for all other types of targets, PCANN algorithm outperforms the CGBC algorithm. For the target type APC also, the performance of PCANN is not consistently worse than the CGBC algorithm. For the sake of conciseness, the comparison between the monostatic and bistatic ATR performance will be done using the PCANN algorithm only.

### 6.2 Monostatic against Bistatic ATR

The comparison between the monostatic and the bistatic ATR is the most important contribution of the present work. Hence, care was taken so that the process of database generation is as similar as possible for the two cases. In the bistatic case, if the transmitter is kept fixed and the receiver is moved round the target, then the bistatic angle goes on increasing, and hence making the product image of increasingly course resolution. To overcome this problem, the (simulated) receiver platform was moved keeping the bistatic angle below a certain angle. For example, in one set of the data collected, the bistatic angle was kept below a rough value of $60°$. In this case, the (simulated) transmitter was kept fixed first at the predecided elevation and at $0°$ azimuth, and the receiver was moved at the same elevation and with the azimuth angle varying from $-60°$ to $+60°$. Next set of data was collected by keeping transmitter azimuth at $120°$ and moving the receiver platform from $60°$ till $180°$. The total three sets of data were collected, so that the final data are again a patch in the $k$-space, with no major gap or jump. Then this patch of $k$-space was used in similar manner as in the monostatic case to form an image clip data base. To see the performance of bistatic ATR for different values of bistatic angles, two sets of bistatic data were collected, one where the bistatic angle is kept less than $30°$ and the other where the bistatic angle is kept less than $60°$.

The ROC curves for the three cases of monostatic, low-bistatic-angle and large-bistatic-angle ATR have been presented in Fig. 5, for all the four different targets.

It can be observed from the ROC curves that the monostatic ATR performance is the best, as expected. But contrary to what might be expected, the bistatic ATR performance is not drastically worse. Another important observation (again contrary to what might have been expected) is that with increasing bistatic angle the ATR performance does not degrade significantly. Performance seems to be more or less the same for the two types of bistatic data used in these experiments. This might be because of the following reason.

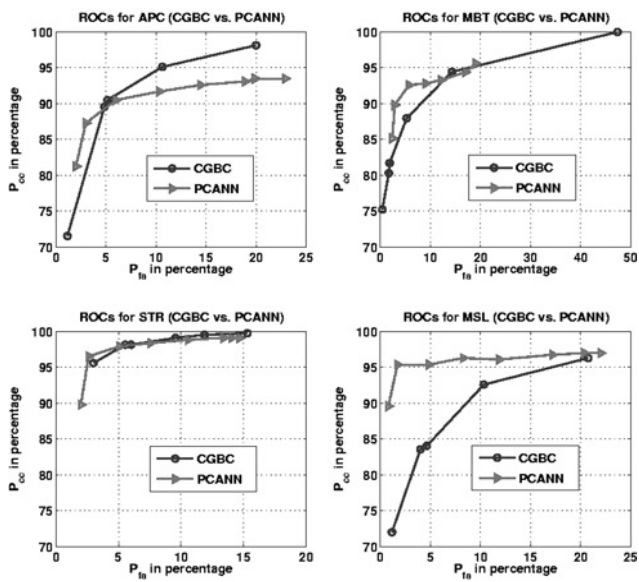

**Fig. 4** *ROC curves for different ATR algorithms*

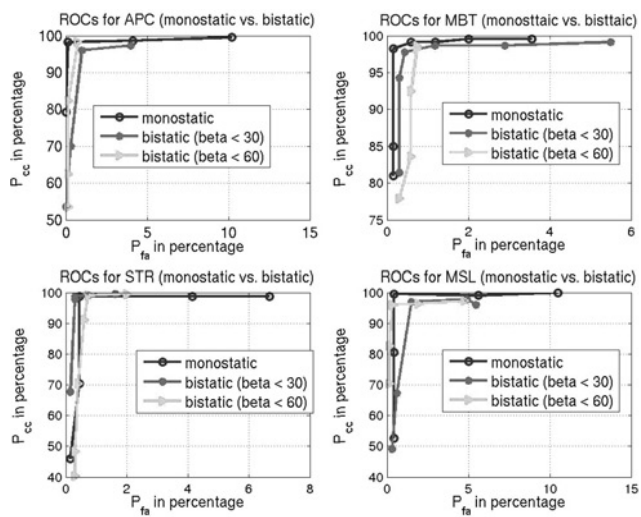

**Fig. 5** *ROC curves for monostatic and bistatic ATR*

In the present work, the target models used are simple and have distinct, major physical features. Hence, the course resolution image clips from even a 60° bistatic angle bistatic configurations may have most of the major features needed for classifying that target type. However, as has been discussed before, the ATR performance should not depend on the detailed features of a target.

### 6.3 Bistatic ATR with increased bistatic angle

With positive conclusions about the abilities of a bistatic ATR system, the next important issue is about the implementation of a bistatic ATR system. It is well known that the image resolution deteriorates with the increase in the bistatic angle of imaging. At very high values of bistatic angle, the image resolution is extremely poor and most of the traditional usages of a SAR-image are no longer possible. Hence, it was deemed pertinent to look into what might be an upper threshold for the value of bistatic angle, for practical purposes. Second, is there any way to make up for the loss of ATR performance, which comes with increasing bistatic angle of imaging?

For this, the data collected were was divided into three sets. The first one consisted of all the image clips formed with the bistatic angle of imaging less than 60°. The second set was formed by those images whose bistatic angle of operation lie between 60° and 80°, and the third set consisted of those images whose bistatic angle of collection lie between 80° and 100°. Both the training and the test datasets were divided in this manner.

The results for ATR experiments in the individual sets using the new PCANN algorithms) is displayed in bar-chart format in Fig. 6.

As can be observed, there is a drop in the performance of the classifier, with increasing bistatic angle of imaging. This can be traced mainly to the fact that with increasing bistatic angle, the frequency ($k$) space support for the image formation step decreases. This in turn reduces the resolution of the images. Hence in normal imaging conditions, images from higher bistatic angle of imaging would be of lower resolution than the same taken with a lower bistatic angle. To make up for this difference, in a separate set of experiments, all the images are formed with same amount of $k$-space support. For this, the $k$-space data for images with lower bistatic angle (in which condition, large $k$-space data are available) were under-utilised. The aim was to use the same amount of $k$-space support for all imaging, as is available for the maximum bistatic angle condition (100° for this case). The results were image-datasets of roughly the same resolution for any bistatic angle.

Graphically, the classification performance in this case is presented in Fig. 7. The drop in the performance with increasing bistatic angle is less severe in this case. The important point to be observed is that if images are formed with similar resolution, then classification performance does not deteriorate much with an increase in the bistatic angle. Some of the important remarks from this are:

1. Bistatic ATR performance does not deteriorate because of the lack of information in bistatic data, rather because of the lack of resolution in the bistatic images.
2. Purely bistatic systems can be used for ATR purpose with sufficient accuracy, and PCANN type classifier can be an algorithm of choice.
3. In practical bistatic systems, if the $k$-space support could be increased with increasing bistatic angle of operation, the ATR performance would not become that sensitive to the bistatic angle of imaging. One of the ways to do this is to use increasing frequency band, for increasing bistatic angle of operation.

### 6.4 Bistatic polarimetric data for ATR

In this subsection, the results from the use of fully polarimetric data for bistatic SAR ATR will be presented. The SAR images from all the polarisations are preprocessed to extract the parameters from the ***K***-matrix, as discussed in Section 5. These parameters in turn are used as features for classification.

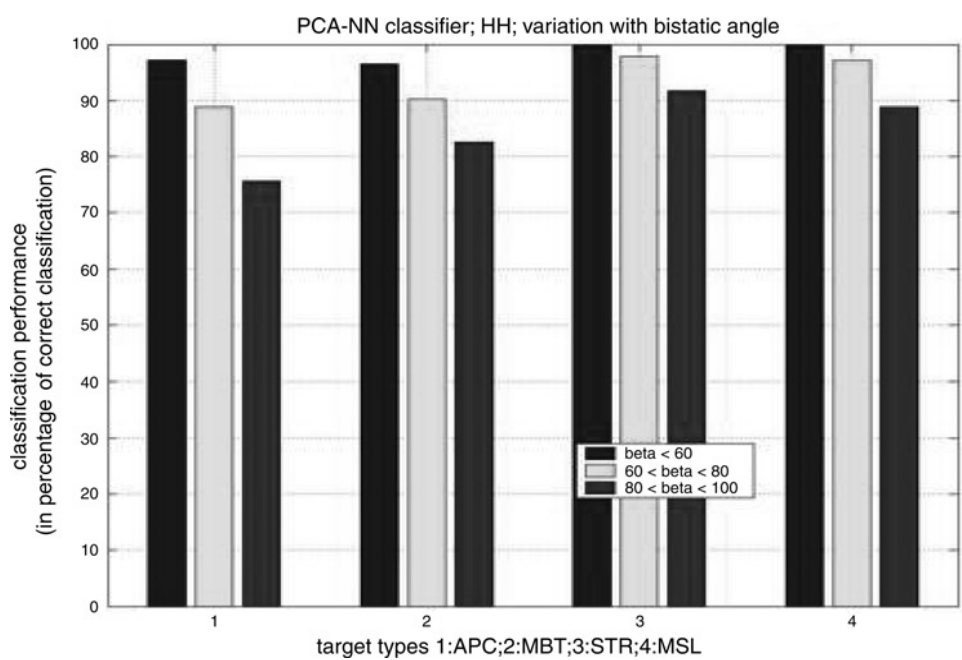

**Fig. 6** *PCANN performance with different bistatic angle*

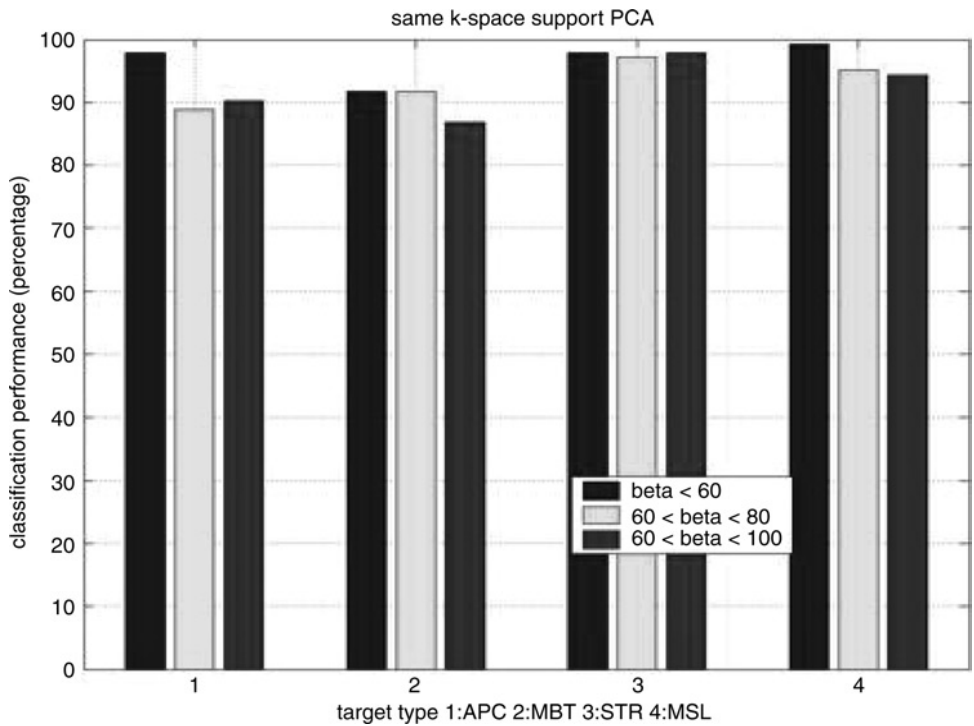

**Fig. 7** *PCA performance with different bistatic angle*
(same *k*-space supported dataset)

Fig. 8 shows the ATR performance for the four targets, as the number of parameters extracted from the ***K***-matrix is increased one by one. The first nine parameters are the ones which are present both for monostatic and bistatic polarimetric cases, and hence are called the monostatic parameters (for convenience). Tenth parameters onwards are the parameters obtainable only for bistatic polarimetric case (when the scattering matrix is not symmetric) and are termed as the bistatic parameters. It can be observed that:

• With increase in parameters, the ATR performance increases.
• With addition of bistatic parameters, the ATR performance does not increase, except for one target (MBT).

The reason why the increase is observed only for MBT can be explained on the basis of the complexity of the models. The MBT is the most complex model among all the four targets modelled in this work. The other three targets are not as complicated as the MBT, and hence the nine monostatic parameters are enough to represent all their classifiable physical features (or the features important for the target classification purpose).

From the above set of results, it can be concluded that the bistatic polarimetric parameters may have information about the targets (which can be used to classify the targets more efficiently). The next experiment was done to find which bistatic parameters, in particular, contain the information. For this the ATR exercise was run with nine of the parameters from the ***K***-matrix. Out of these, eight were kept fixed. For the ninth parameter, all the bistatic parameters were used one by one. The results are shown in Fig. 9.

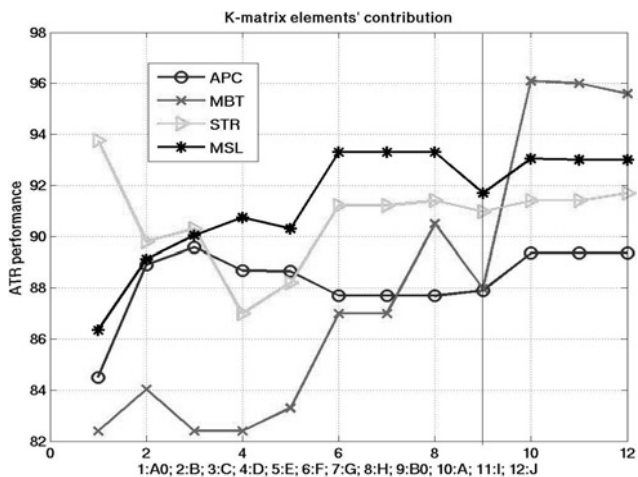

**Fig. 8** *ATR performance with increased number of parameters derived from the **K** matrix*

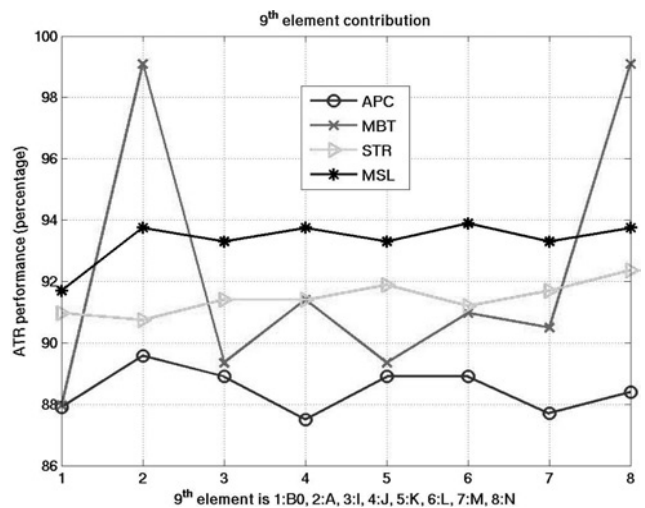

**Fig. 9** *Ninth feature effect on ATR*

As before, the performance increases only for the target MBT. The interesting observation from this experiment is that only two out of all the extra bistatic parameters used are responsible for the increase in ATR-performance. This shows that those two parameters have the extra information responsible for the increase in the ATR performance with the inclusion of the bistatic parameters (derived from the **K**-matrix). Hence, it is very likely that these two parameters have some strong correlation to certain physical features of the target (similar to the type of correlation drawn in the monostatic case between the physical features of a target and the parameters derived from the **K**-matrix [6].

## 7 Limitations of the study

In this section some of the limitations of the present work will be discussed. The major limitation of the work is the non-availability of real SAR images for validating the ATR exercises. As an alternative, simulated database was used in the study. In simulating the database, EM simulators were used. Because of computational load, the target models were kept simple. The same simplistic models were used to generate both the training and test datasets. In reality the targets are much more detailed causing the EM return to consist of many higher-order scattering phenomena. However, most of these detailed EM scattering return from targets are common to all types of vehicles and hence do not act to distinguish a target from another. Second, almost all the ATR algorithms presented in the present paper have been tested using the real MSTAR SAR data available in the public domain [12]. It has been observed that testing the algorithms using real database showed similar trends as using simulated database, that is, algorithms' relative performances remained the same. Hence, even though the absolute figures found in the present study may not hold true when handling the real data, the relative performances and conclusions are expected to hold true irrespective of the simulated nature of the dataset.

Another limitation of the work is the clutter and other noise components in the simulated dataset. Because of the simplistic models used, the images contain very little clutter and noise from the higher-order EM scattering phenomena. Hence, the use of PCANN was more effective in this case. However, as reported in one of our studies [7], the use of PCANN on real monostatic MSTAR dataset shows similar results and also establishes the superiority of the PCANN algorithm as compared with many of the established TAR algorithms.

Third, it is easier in the current work to extract the polarimetric features from the dataset, as the data were simulated and hence highly controlled. However, collecting such pure multipolar data in reality is an extremely difficult task as of now. It is expected that with newer methods of calibrating polarimetric radar systems, collecting pure multipolar data will be feasible. In the current work, the main intention was to show the information content of the bistatic multipolar data, and the simulated datasets have helped in establishing it within the present limitations.

In studying the ATR algorithms, no confuser class could be included in the test dataset. This was primarily because of the time taken to simulate the dataset for each of the target types. Under this limitation, to test the confidence measure of the ATR algorithms the ROC curves are used instead. In generating the ROC curves for a target, the risk of misclassifying that target is varied and the ATR performance was measured. This gives a limited measure of the confidence of the ATR algorithms.

The centre frequency used in the simulations was 1-GHz and the bandwidth of simulation was 450 MHz. This makes the system that of an ultra wide bandwidth. This choice of bandwidth and centre frequency was dictated by the limitations of the EM modelling tool, the physical size of the targets and the computing resources available. This sets a lower limit on the wavelength (and upper limit on the frequency) that was possible to use in simulation. The bandwidth was then chosen to provide similar resolution as of conventional X-band SAR systems.

## 8 Conclusions

In this limited investigation on bistatic ATR, the major conclusion is that, contrary to popular reservations, ATR can be implemented using bistatic systems without much loss in the performance, as compared with the existing monostatic ATR systems. Second, again contrary to certain existing reservations, the bistatic ATR performance does not deteriorate much, with an increase in the bistatic angle of operation. By using increasing bandwidth, the loss of ATR performance suffered with increasing bistatic angle can be made up for (for bistatic angles of as high as $100°$). Hence, the loss in the ATR performance with increasing bistatic angle is with all likeliness not because of any loss in the information content, but because of the loss of image resolution in the bistatic configuration. Another finding from the current work is that, in contrast to comments reported elsewhere, the bistatic polarimetric data do contain additional information about the target and possibly represent certain physical features of the target, like the monostatic multipolar data. Finally, the new ATR algorithm developed in the present work, viz., the PCANN algorithm is efficient for bistatic ATR and is computationally extremely fast.

The lack of real and field collected data was the major bottle neck for this work. However, simulated data were quite reliable to make a qualitative conclusion on ATR using bistatic radar systems. With the present work, the arguments for implementing bistatic systems are expected to gain more support.

## 9 Acknowledgment and disclaimer

The authors acknowledge the Electromagnetic Remote Sensing Defence Technology Centre (EMRSDTC) for funding this project and the Royal Academy of Engineers (RAE) for an equipment grant. Any views expressed are those of the authors and do not necessarily represent those of the MOD or any other UK government department.

## 11 Appendix

In this section the eigenvalue analysis of the radar image data is presented to show that within standard SAR imaging assumptions, principal components represent the scattering centres from a SAR image. In one of the previous work [19], the PCA relevance of the scattering centres is shown by analysing the real SAR images of military targets. In another work [7], PCANN is applied to ATR using real SAR image database [12]. These two previous work show that in spite of the noise present in SAR images, PCA is effective in extracting the scattering centre information from the SAR images.

Traditionally, radar imaging has depended on the scattering centre assumption. According to this, if wavelength of the illuminating EM wave is smaller than the object dimensions, the scattered return could be modelled as coming from distinct scattering centres. This model when applied to radar imaging gives the scattering centre modelling of the radar images, where the radar image is modelled to be the summation of distinct scattering centres. The theoretical representation of an ideal scattering centre is the two-dimensional Kronecker's delta function. Practically, more approximate mathematical functions can be used. In the present analysis, a two-dimensional sinc function is taken as the model for an ideal scattering centre in a radar image. Let us analyse a simple radar image, with a single scattering centre in the scene, at the centre of the scene. Let the size of the scene be $N \times N$ pixels. The image can be represented by a matrix $A$ of pixel amplitudes, which according to the scattering centre model can be modelled with a two dimensional sinc function.

$$A = \bar{u} S \bar{v}^{H} \qquad (12)$$

Here, $\bar{u}$ and $\bar{v}$ are two one-dimensional sinc vectors (vectors of dimension $N \times 1$) and $S$ a diagonal matrix with the diagonal set to the value one (Hence $S$ is unit matrix), and of dimension $1 \times 1$. $()^{H}$ represents the Hermitian operation on a matrix. Two of the important characteristics of a scattering centre are its amplitude and position. It can be observed from the above equation that:

- To change the intensity of the scattering centre, the value of the diagonal elements of the $S$ matrix needs to be changed.
- To shift the position of the scattering centre, the sinc vectors, viz., $\bar{u}$ and $\bar{v}$ are to be shifted appropriately.

Similarly, a scene with $k$ scattering centres at different positions can be represented as:

$$A = USV^{H} \qquad (13)$$

Here (Dimension of $A$ is $N \times N$, of $U$ is $N \times k$, of $S$ is $k \times k$, and of $V^{H}$ is $k \times N$.) $U = [\bar{u}_1 \ \bar{u}_2 \ \bar{u}_3 \ \ldots \ \bar{u}_k]$, and $V = [\bar{v}_1 \ \bar{v}_2 \ \bar{v}_3 \ \ldots \ \bar{v}_k]$. Each element vectors $\bar{u}_i$ and $\bar{v}_i$ are sinc-function vectors of length $N$, and appropriately shifted to represent the position of the $i$th scattering centre. $S$ is a diagonal matrix of size $k \times k$, with the $i$th diagonal element representing the intensity of the $i$th scattering centre.

Because the vectors $\bar{u}_i$'s and $\bar{v}_i$'s are sinc functions, the matrices $U$ and $V$ are unitary and orthonormal. This conclusion holds true for most of the mathematical functions which could be taken to model an ideal scattering centre. Given these facts, (12) is in the same form as the singular value decomposition (SVD) of the image pixel matrix. Hence, it can be observed that radar images can be decomposed as per SVD (provided the scattering centre assumption is true). Few points worth noting are:

- The number of scattering centres determines the number of elements in $S$, and hence the rank of the final image matrix.
- The position of the scattering centres is determined by the elements of $U$ and $V$ matrices.
- The strength of the scattering centres is determined by the singular values (i.e. The elements of the diagonal matrix $S$)
- The $i$th element of $S$ can be shown to be equal to $\lambda_i^{1/2}$, where $\lambda_i$ is the $i$th eigenvalue of the covariance matrix of $AA^{H}$ [18, 25].
- Columns of $U$ are the eigenvectors of $AA^{H}$, and those of $V$ are the eigenvectors of $A^{H}A$ [17, 26]. Let it be assumed that in the image matrix $A$, the rows represent range and columns represent cross-range vectors. Then $AA^{H}$ is the covariance matrix (assuming the image has been zero-centred) of the cross-range pixel variables, and $A^{H}A$ represents the covariance matrix (assuming the image has been zero-centred) of the range pixel variables. Hence, pre-multiplication of the image matrix with $U^H$ represents the PCA with respect to cross-range pixels and multiplication of the image matrix with $V$ represents PCA with respect to range pixels [18].

$$U^H A V = U^H (U S V^H) V \quad (14)$$
$$= I S I = S \quad (15)$$

- The second step can be explained as both $U$ and $V$ are unitary, that is, $U^H = U^{-1}$ and $V^H = V^{-1}$, $U_H U = V_H V = I$. Here $I$ is the identity matrix and $U^{-1}$ represents the inverse of the matrix $U$. Hence, applying PCA in one dimension is equivalent to extracting the position of the scattering centres in that dimension. Hence the result of applying PCA is a two-dimensional matrix with information about the scattering centre intensities.
- In real images, there is the presence of noise. Hence, the choice of the dimension of $S$, $U$ and $V$ depends on the choice of $k$, the number of the largest eigenvalues of $AA_H$ to be taken. This is similar to the super-resolution algorithms [25], which have been applied by some researchers in filtering the noise from the radar images [27, 28].
- When PCA is performed on a radar image, matrices $U$ and $V$ are calculated. These matrices not only help to reduce the dimension of the matrix $D$, but represent the position of the scattering centres in the image matrix $A$.
- The eigenvectors chosen in forming $U$ and $V$ correspond to the largest eigenvalues, and hence correspond to the brightest scattering centres in the image matrix $A$.
- Hence, applying PCA is equivalent to extracting the position and intensity information of the scattering centres.

*Use in classification*: In training phase, given the images of the target of a particular class, the $U$ and $V$ matrices are found out,

- which in turn correspond to the positions of the scattering centres in the target image, and
- which, when applied on an image of the target, will give a matrix $S$, which will be a diagonal matrix with elements in diagonal, corresponding to the intensity of the brightest scattering centres.

Given a test image, applying $U$ and $V$ on that and finding the distance (Euclidean distance for simplicity) of the resulting matrix from $S$ in essence represents the task of comparing the position and intensities of the major scattering centres in the test image with the target class with whose training the matrices $S$, $U$ and $V$ have been found.

The above analysis shows the correlation between PCA of radar images and scattering centre analysis of the radar images. However, the analysis is limited in a few aspects. First, it does not hold whether the scattering centre model of the radar image does not hold correct [29]. Second, here the two-dimensional PCA has been presented, where PCA is applied both to the range and to the cross-range dimensions of the radar image. In the current project, a stacking operation is performed on the radar images to form one-dimensional vectors, and then one-dimensional PCA is applied. However, the stacking operation is linear. Hence, the results derived in the current analysis would still hold true. Moreover, it has been checked by applying the two- and one-dimensional PCA to radar images that both ways of applying PCA result in the same ATR performance [21].